\documentclass{article}

\PassOptionsToPackage{numbers, compress}{natbib}

\usepackage[preprint]{neurips_2019}

\usepackage[utf8]{inputenc} 
\usepackage[T1]{fontenc}    
\usepackage{hyperref}       
\usepackage{url}            
\usepackage{booktabs}       
\usepackage{amsfonts}       
\usepackage{nicefrac}       
\usepackage{microtype}      
\usepackage[pdftex]{graphicx}
\usepackage{subfigure}

\usepackage{footnote}
\makesavenoteenv{tabular}
\makesavenoteenv{table}

\title{Scale MLPerf-0.6 models on Google TPU-v3 Pods}

\author{
Sameer Kumar, Victor Bittorf, Dehao Chen, Chiachen Chou, Blake Hechtman, HyoukJoong Lee, \\
\bf{Naveen Kumar, Peter Mattson, Shibo Wang, Tao Wang, Yuanzhong Xu, Zongwei Zhou} \\
Google Research, Brain Team~\\
\texttt{\{sameerkm, vbittorf, dehao\}@google.com}~\\
}
\date{September 6 2019}

\begin{document}

\maketitle

\begin{abstract}
The recent submission of Google TPU-v3 Pods to the industry wide
MLPerf v0.6 training benchmark demonstrates the scalability of a suite
of industry relevant ML models. MLPerf defines a suite of models,
datasets and rules to follow when benchmarking to ensure results are
comparable across hardware, frameworks and companies. Using this suite
of models, we discuss the optimizations and techniques including
choice of optimizer, spatial partitioning and weight update sharding
necessary to scale to 1024 TPU chips. Furthermore, we identify
properties of models that make scaling them challenging, such as
limited data parallelism and unscaled weights. These optimizations
contribute to record performance in transformer, Resnet-50 and SSD in
the Google MLPerf-0.6 submission.

\end{abstract}

\section{Introduction}

MLPerf \cite{mlperf} is a machine learning benchmark suite that has
gained industry wide support and recognition. Recently, in Jul 2019,
the second round of results for the training benchmarks, MLPerf v0.6,
were published including submissions from NVIDIA, Intel, Google,
Fujitsu and Alibaba. Submissions ranged in size from a machine with 8
accelerators to clusters with over 1000 accelerators using ML
frameworks including Tensorflow, Pytorch and MXNet and
others.~\footnote{MLPerf also benchmarks ML inference performance and
  the first inference submission is expected in late 2019.}  Like
systems benchmark suites which have come before it, the MLPerf
Benchmark suite is pushing performance forward and our v0.6 MLPerf
submission on Google TPU-v3 accelerators showcases the large scale we are able
to achieve. MLPerf follows in the footsteps of SPEC
\cite{DBLP:journals/pc/Dixit91} and TPCH \cite{tpch} to create an
industry standard benchmark suite for ML systems including
accelerators, frameworks and modeling on state of the art ML training
tasks. Not only does MLPerf allow for comparisons across frameworks
and hardware, but it fundamentally drives understanding and
development of ML systems and methodology.
	
An MLPerf training benchmark involves training a model
(e.g. Resnet-50) on a specific dataset (e.g. Imagenet) while following
specific methodology for parameters, optimizations, and timing. For
v0.6, the MLPerf rules were expanded to enable larger scale of systems
to submit to the benchmark. Particular changes included allowing the
LARS optimizer for Resnet-50 and a time budget allowing for large
scale systems to initialize while also increasing the accuracy
requirements for the trained models. MLPerf is still challenging to
run at scale, for example the rules require implementations to context
switch between training and evaluation every few seconds at large
scales which incurs significant overhead not seen in production use
cases. MLPERF-0.6 accuracy targets present a significant challenge at
scale as increasing the global batch size can reduce the accuracy that
can be achieved.

\begin{figure}
    \centering
    \begin{minipage}{0.475\textwidth}
        \centering
        \includegraphics[width=1.0\textwidth]{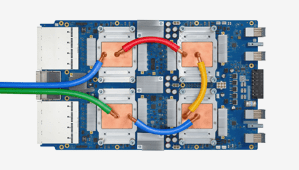}
        \caption{Google TPUv3 device with four chips, 420 teraFLOPS and 128 GB of HBM.}
        \label{fig:tpuv3}
    \end{minipage}\hfill
    \begin{minipage}{0.475\textwidth}
        \centering
        \includegraphics[width=1.0\textwidth]{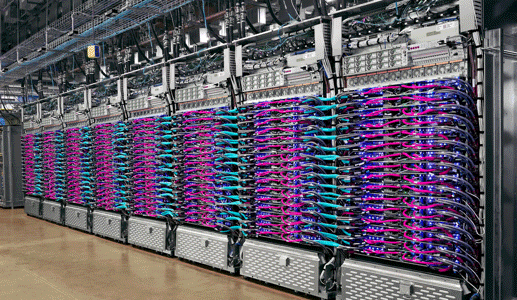}
        \caption{Google TPU-v3 pod with 1024 chips, 107 PetaFlops and 32 TB of HBM interconnected by a 2-D torus network.}
        \label{fig:tpupod}
    \end{minipage}
\end{figure}

In this paper, we present techniques used to optimize MLPerf benchmark
results on the third generation Google Tensor Processing Units
(TPU-v3) shown in Figure~\ref{fig:tpuv3}.  The Google
TPU-v3 is an ML accelerator designed to accelerate neural network
workloads by enabling significant matrix-matrix and matrix-vector
compute acceleration on each TPU-v3 chip coupled with 32 GB of high bandwidth
memory and 32 MB of scratchpad memory for storing weights and
activations, respectively.  Each TPU chip has two separate cores.
In a TPU-v3 pod (Figure~\ref{fig:tpupod}), 1024 TPU-v3 chips are interconnected 
by a custom high throughput 2-D torus interconnect to accelerate remote DMA and global
summation operations. 


\section{Methods}

We present performance optimization techniques to optimize MLPerf 0.6
training time on TPU-v3 pods.  We use ~\cite{tensorflow2015-whitepaper,dean2012large} for all the MLPerf
0.6 benchmarks. The TensorFlow graphs are lowered by the XLA compiler
\cite{xla} to the cloud TPU-v3 pods. The XLA compiler enables various
optimizations like unrolling and pipelining loops and fusion of
compute kernels to maximize the execution throughput of the matrix
unit~\cite{Jouppi_17} on cloud TPU-v3 accelerator cores. We use mixed
precision with the bfloat16 precision in all our benchmark runs
\cite{bf16}. To maintain comparable accuracy with 32-bit floating
point networks, all non-convolutional operations (e.g. batch
normalization, loss computation, gradient summation) use 32-bit
floating point numbers. Since the majority of the computational and
memory access overheads in MLPerf models are in the convolutional
operations, use of bfloat16 enables higher training throughput with
minimal or no loss in model accuracy.  When the number of examples per
TPU accelerator is below a threshold, we use the distributed
normalization technique presented in \cite{resnet_18}.  The TensorFlow
runtime on TPU-v3 pods execute the input pipeline to pre-process
inputs on host CPUs. We use caching, host to device offload of select
TF ops and prefetching \cite{resnet_18} to optimize the host input
pipeline throughput. In addition, we explore the following
optimization techniques to achieve peak scaling on TPU-v3 pods.


{\bf Distribute evaluation computation}: in a traditional TensorFlow
model trained on a cloud TPU-v3 pod, the evaluation job is executed
separately on a side card with additional TPU chips. In the MLPerf
models, the execution of the evaluation metric can become an Amdahl
bottleneck limiting the scalability of the benchmark.  We designed a
new train and evaluation tight loop that is executed on the TPU
accelerators. Both train and evaluation are distributed on all the
TPU-v3 pod accelerator cores. The output evaluation metric tensor is
computed at the epochs specified in the MLPerf rules.  For example, in
ResNet-50, the eval metric tensors are computed every 4 epochs. The
evaluation metric tensors are used to compute top-1 accuracy published
in the training job’s standard output.  The evaluation dataset is
padded with zeros when the evaluation examples is not a multiple of
the evaluation batch size. Only output tensors from the TPU cores that
have real examples is considered while computing the top-1 accuracy
metric.


{\bf Optimize gradient summation}: we use the 2-D gradient summation
technique presented in \cite{resnet_18} to aggregate gradients on the
TPU-v3 torus network. We observed MLPerf TensorFlow benchmarks with
non-contiguous gradient tensors had limited gradient summation
throughput. We optimized the 2-D scheme by pipelining gathers from
non-contiguous tensors from HBM to on device memory with summation of
network packets in the reduction operation. In the broadcast phase the
scatters of the result buffers to non-contiguous storage is pipelined
with data transfer on the network. This aggressive pipelining of the
gradient summation results in over 1.5x speedup of gradient summation
throughput in the ResNet-50 model on TPU-v3 pods.

\begin{figure}[t]
    \centering
    \begin{minipage}{0.475\textwidth}
        \centering
        \includegraphics[width=1.0\textwidth]{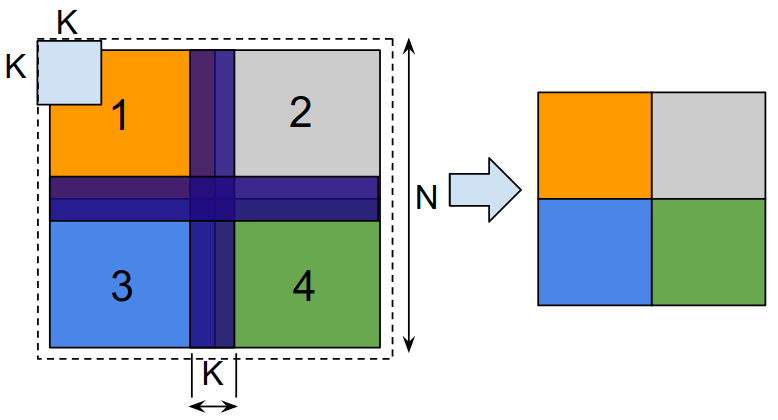}
        \caption{Spatial partitioning of a 2-D convolution with an NxN input and kernel size K on 4 cores.}
        \label{fig:spatial}
    \end{minipage}\hfill
    \begin{minipage}{0.475\textwidth}
        \centering
        \includegraphics[width=1.0\textwidth]{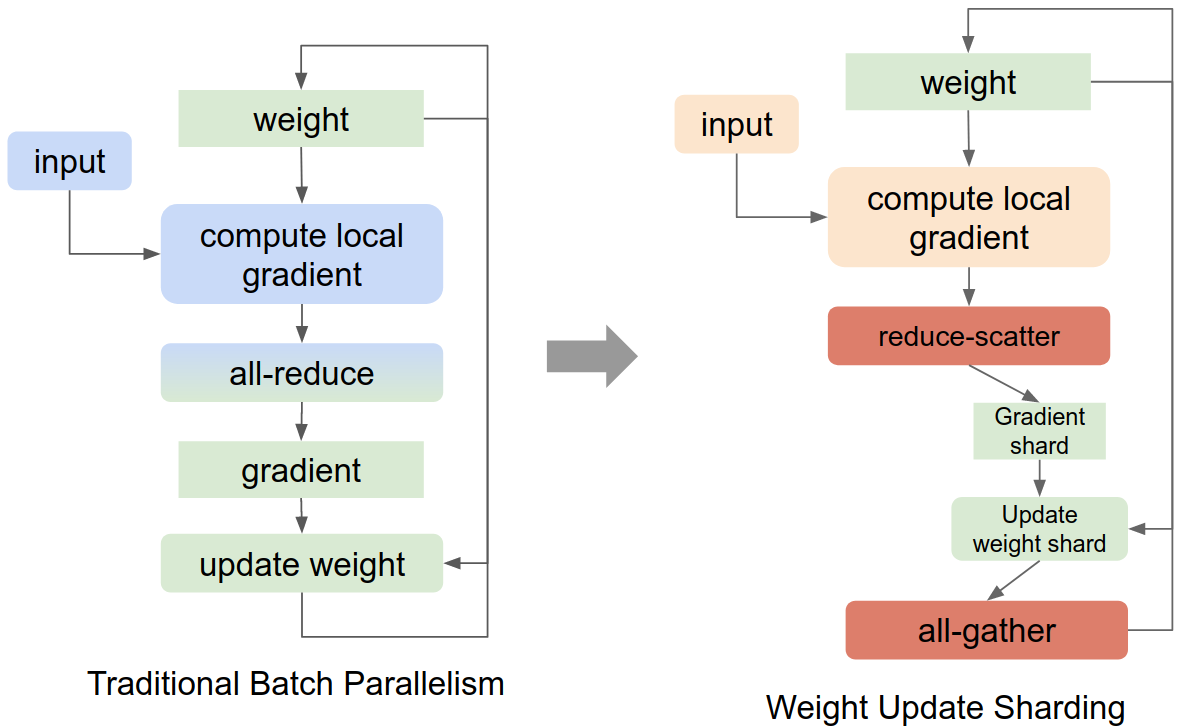}
        \caption{Weight update sharding on TPUv3 pods}
        \label{fig:weight_update}
    \end{minipage}
\end{figure}


{\bf Model parallelism}: as the batch sizes are small in some of the
MLPerf models, we use model parallelism to enable higher parallelism
in those benchmarks. We use the following two model parallelism
techniques to achieve higher scaling in the MLPerf benchmarks:

\begin{itemize}
\item Spatial Partitioning. In this technique MLPerf computation
  kernels are partitioned along both batch and spatial dimensions to
  increase parallelism and enable execution on a larger number of
  TPU-v3 accelerator cores. Halo exchange communication operations are
  added to synchronize TPU-v3 cores that execute spatially partitioned
  workloads (Figure~\ref{fig:spatial}).
    
\item Weight update sharding.  When the number of examples per TPU-v3
  accelerator core is small, we observe the optimizer weight update
  computation results in significant overheads. For example, with
  ResNet-50 on 2048 TPU-v3 cores, the LARS optimizer weight update
  overhead is about 6\% of the total device step time. In the MLPerf
  Transformer model, the ADAM optimizer weight update time is about
  45\% of the step time. So, we distribute the weight update
  computation across TPU-v3 cores, and then use an optimized
  all-gather to broadcast the new weights to all the TPU-v3 cores
  (Figure~\ref{fig:weight_update}).
\end{itemize}

\section{Benchmark Analysis}
In this section, we present case studies for five MLPerf-0.6
benchmarks. In addition to the techniques presented above, we also
explore specialized optimizations for these MLPerf models.


{\bf ResNet-50}: MLPerf uses the ResNet-50 model
\cite{DBLP:journals/corr/HeZRS15} on the ImageNet-1K \cite{ILSVRC15}
dataset to benchmark image classification. ResNet-50 is one of the
most widely used models for benchmark ML and MLPerf uses a specific
variant of ResNet-50 termed "version
1.5"~\cite{DBLP:journals/corr/GoyalDGNWKTJH17} to indicate a slight
modification to the model architecture from the original which is
commonly found in practice. In order to scale the ResNet-50 MLPerf
benchmark to the 2048 core TPU-v3 pod system, we used batch
parallelism along with the distributed evaluation, distributed batch
normalization, weight update sharding and gradient summation
optimizations.


\begin{figure}[b]
    \centering
    \begin{minipage}{0.45\textwidth}
        \[\lambda = \epsilon \times ||w|| / (||g|| + \beta \times ||w||)\]
        \[v = m \times v + (g + \beta \times w)\]
        \[w = w - \lambda \times v\]
        \caption{Scaled momentum \cite{mlperf}.}
        \label{eqn:tf_ref}
    \end{minipage}\hfill
    \begin{minipage}{0.5\textwidth}
        \[\lambda = \epsilon \times ||w|| / (||g|| + \beta \times ||w||)\]
        \[v = m \times v + \lambda \times (g + \beta \times w)\]
        \[w = w - v\]
        \caption{Unscaled momentum \cite{lars_32k}.}
        \label{eqn:lars_paper}
    \end{minipage}
\end{figure}

The MLPerf-0.6 reference for Resnet-50 uses the adaptive learning rate
scaling LARS optimizer~\cite{lars_32k}. It enables training to target
accuracy in 72 epochs at batch size 32768. The reference LARS
optimizer uses the weight update equation shown in
Figure~\ref{eqn:tf_ref}. Here, $\lambda$ is the learing rate, $g$ is
the gradient tensor, $w$ is the weight tensor, $\beta$ is the weight
decay, $m$ is the momentum hyper parameter and $\epsilon$ is the LARS
coefficient. This LARS optimizer presented in
literature~\cite{lars_32k} uses a weight update equation shown in
Figure~\ref{eqn:lars_paper}.  Notable difference is that the momentum
parameter is scaled by the learning rate in the MLPerf reference.  A
systematic study of the LARS optimizer is beyond the scope of this
paper. However, we find the MLPerf ResNet-50 model converges in 70.6
epochs via the optimizer update equation shown in
Figure~\ref{eqn:lars_paper}. Further, tuning the momentum
hyper-parameter enables training in only 64 epochs with a {\bf record
  benchmark time of 67.1 seconds}. Table 1 summarizes the benchmark
times for the MLPerf-0.6 Resnet-50 experiments. Note, tuning the
momentum parameter is not permitted by the MLPerf-0.6 submission rules
in the closed division category.

\begin{table}
  \centering
  \begin{tabular}{lccccl}
    \toprule                \\
    Optimizer & Base LR & Warmup Epochs & Momentum & Train Epochs & Benchmark \\
    \midrule
    Scaled momentum & 31.2 & 25 & 0.9 & 72.8 & 76.9~\footnote{Google MLPerf-0.6 Submission.} \\
    Unscaled momentum & 31.2 & 25 & 0.9 & 70.6 & 72.4 \\
    Unscaled momentum & 29.0 & 18 & 0.929 & 64 & 67.1 \\
    \bottomrule
  \end{tabular}
   \caption{ResNet-50 benchmark seconds on 2048 TPU cores and batch 32K.}
  \label{tab:resnet_pod}
\end{table}  


{\bf SSD:} Single Shot Detection~\cite{ssd_15} is one of two object
detection models in the MLPerf benchmark; SSD is intended to reflect a
simpler and lower latency model for interactive use cases such as in
end-point and non-server situations. Notably, SSD uses a pre-trained
ResNet-34 backbone as part of the architecture. SSD is trained and
evaluated on the COCO dataset~\cite{microsoft_coco}.

Note the computational overhead of the SSD model is small compared
with the ResNet-50 model. So, we explore both data and model
parallelism to scale SSD to TPU-v3 pods. We use spatial partitioning
to parallelize SSD on up to 4 TPU accelerator cores. Achieving high
speedup from spatial partitioning is challenging due to the following:
\begin{itemize}
\item Higher communication overheads: spatial partitioning results in communication overheads from halo exchange between spatial partitioned neighbors. In addition, it results in all-reduce calls for distributed batch normalization executed on large number of workers. 
\item Load imbalance: In our current XLA implementation of spatial partitioning, some TF operations are not sharded and executed on spatial worker 0 resulting in a load-imbalance. 
\item Relatively small spatial dimensions: The spatial dimensions in SSD is decreased from 300x300 in the first layer to 1x1 in the last layer. The deeper layers of SSD have smaller spatial dimensions and larger feature dimensions. This results in limited parallelism from spatial partitioning of the deeper layers. 
\end{itemize}


{\bf Mask-RCNN}~\cite{DBLP:journals/corr/HeGDG17} is the more complex
of the two object detection benchmarks in MLPerf. Besides object
detection, Mask-RCNN also performs instance segmentation, which
assigns a semantic label as well as an instance index to each pixel in
the image. Unlike SSD, which is a one stage detector, Mask-RCNN has
two stages: one for proposing instance candidates and the other for
fine-tuning the proposals. Also, Mask-RCNN uses a larger image size
than SSD even though they both train in the COCO dataset. Furthermore,
Mask-RCNN uses a Resnet-50 backbone plus Feature Pyramid Network
contrasted with SSD's use of Resnet-34. Scaling Mask-RCNN is
particularly challenging as this model did not converge to the target
evaluation accuracy on a global batch size larger than 128. This
prevents Mask-RCNN from scaling to a large number of cores beyond 128
by just reducing per-core batch size. We use a combination of data and
model parallelism to scale Mask-RCNN beyond 64 TPU cores. We use
spatial partitioning to to parallelize the first stage of
Mask-RCNN. In the second stage, we apply graph partitioning by placing
independent ops on up to four different cores.




{\bf Transformer}~\cite{DBLP:journals/corr/VaswaniSPUJGKP17}
represents state-of-the-art language translation in the MLPerf suite
and is one of two translation models. Trained on the WMT English to
German dataset~\cite{wmt_17}, Transformer uses an attention-based model which
differentiates it from the other language model in MLPerf, GNMT.

To scale Transformer to a full TPU-v3 pod, we used data parallelism
along with the distributed and in-memory evaluation, weight update
sharding, and gradient summation optimizations. We use a global batch
size of 2048 (batch 1 per core), that is dramatically higher than the
reference default batch size. To enable large batch training~\cite{DBLP:journals/corr/KeskarMNST16}, we tuned
hyper parameters to reduce the number of epochs to convergence. We
found increasing the learning rate and tuning warmup steps
insufficient to train the transformer model with a large batch size.
In addition, the beta1 and beta2 hyper parameters of the Adam
optimizer had to be tuned along with a lower learning rate to converge
the MLPerf Transformer model to the target accuracy.

As transformers typically have attention layers that are large fully
connected layers, they have significantly higher number of parameter
weights. Moreover, the overhead of weight updates in distributed
training is significant. The weight update sharding technique in the
XLA compiler solves this by reducing the overhead weight update
operation. The fast 2-D gradient summation technique optimizes
gradient aggregation throughput on the TPU-v3 pods.

As the training time becomes smaller on large TPU pod slices, we
observed the eval and infrastructure overheads dominate the end-to-end
convergence time. To reduce infrastructure overheads, distributed and
in-memory evaluation and nested train-and-eval loop techniques are
adopted. Further, redundant gather operations are removed from the
model. Bfloat16 mixed precision is used to reduce the memory pressure
from matrix multiplication operations. In addition, the maximum
sequence length is reduced from 256 to 97 to reduce evaluation
overheads on TPU cores. Note, 97 is the length of the largest example
in the evaluation dataset.


{\bf GNMT}~\cite{GNMT_16} is the other language translation benchmarks
in MLPerf that is differentiated by its use of recurrent neural
network (RNN). While GNMT achieves a lower target accuracy than
Transformer, the use of a RNN may allow the performance insights
to other RNN models that are generally used by machine learning
community. Like Transformer, GNMT uses WMT English to German for
training and evaluation.

The most expensive computation of GNMT is the gate function
computation in the cell function of the RNN loop. GNMT uses standard
LSTM cells, which concatenate the input feature and the hidden state
of the previous step, and perform dot-product on the concatenated
feature to produce the 4096 output features. For the first
uni-directional layer in encoder, the output of the bidirectional
layers are concatenated to form the input. For the decoder layers,
attention feature is also concatenated with the previous layer’s
output to form the input.


Each RNN layer iterates until all sequence non-padded tokens have been
processed with the entire batch. Because of synchronous training, each
training step will wait until the longest sequence to finish before
the gradient can be accumulated across all workers. To achieve good
load-balance, we use a window based bucketization scheme to ensure
that the sequences in each batch have similar length. For multi-host
training, global bucketization is enabled by using a single host to
produce the input for all workers. This is only possible because the
GNMT inputs are small and preprocessing is inexpensive. However, when
scaling to very large systems where we have 1024 workers, the single
host input pipeline becomes the bottleneck. We use a round-robin
algorithm to distribute the input pipeline to multiple hosts to 
parallelize the workload while maintaining good load balance.




When the per-core batch\_size is small, the LSTM cell computation is
memory bound. As the largest converging global batch\_size
is fixed, per-core batch\_size is small on a
large scale system. Minimizing the input\_feature is an
effective solution to reduce the memory bandwidth requirements for
this model. In an LSTM based RNN loop, the previous step's hidden
state is the next step's input to form a loop carried dependency. But
the projection on the input feature can happen in parallel. So we
hoisted the input feature projection out of the RNN loop
so that we can process many step’s input features in parallel to
maximize the effective batch size. Inside the RNN loop, we only do
projection on hidden state, the output of which is added to the
projected input to derive the output. This optimization is
mathematically equivalent with the traditional LSTM, but much more
efficient for small per-core batch\_size. For the backward path, we
do similar optimization to move the gradient computation part out of
the RNN loop. Instead of computing gradient for every time step and
accumulate it inside the loop, we save the input to an array of full
time range and only update this array inside the RNN loop. After the
RNN loop finishes, we compute the accumulated gradients all at once to
maximize the effective batch size.

\section{Results}

\begin{figure}
    \centering
    \begin{minipage}{0.475\textwidth}
        \centering
        \includegraphics[width=1.0\textwidth]{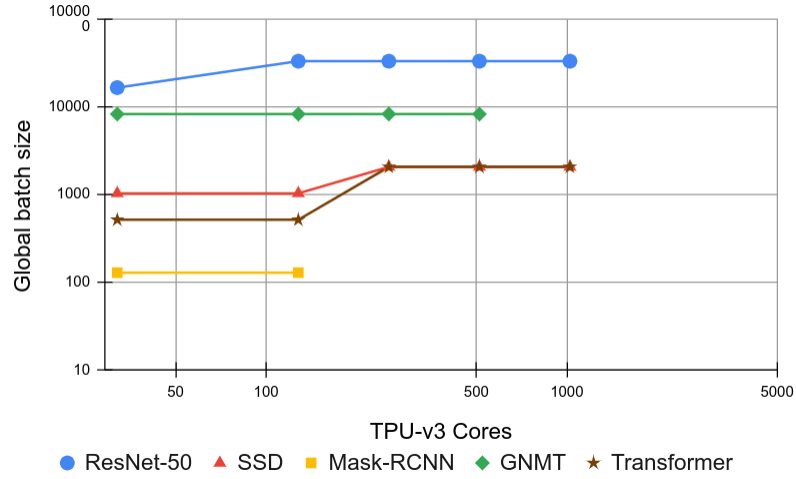}
        \caption{Batch sizes used in scaling MLPerf models.}
        \label{fig:batch_size}
    \end{minipage}\hfill
    \begin{minipage}{0.475\textwidth}
        \centering
        \includegraphics[width=1.0\textwidth]{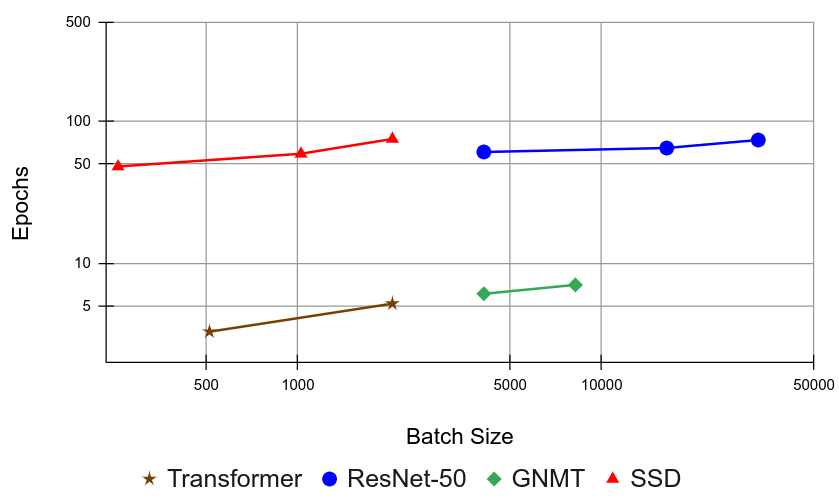}
        \caption{Training epochs to converge when scaling to a larger batch size.}
        \label{fig:epochs}
    \end{minipage}
\end{figure}

\begin{figure}
    \centering
    \begin{minipage}{0.475\textwidth}
        \centering
        \includegraphics[width=1.0\textwidth]{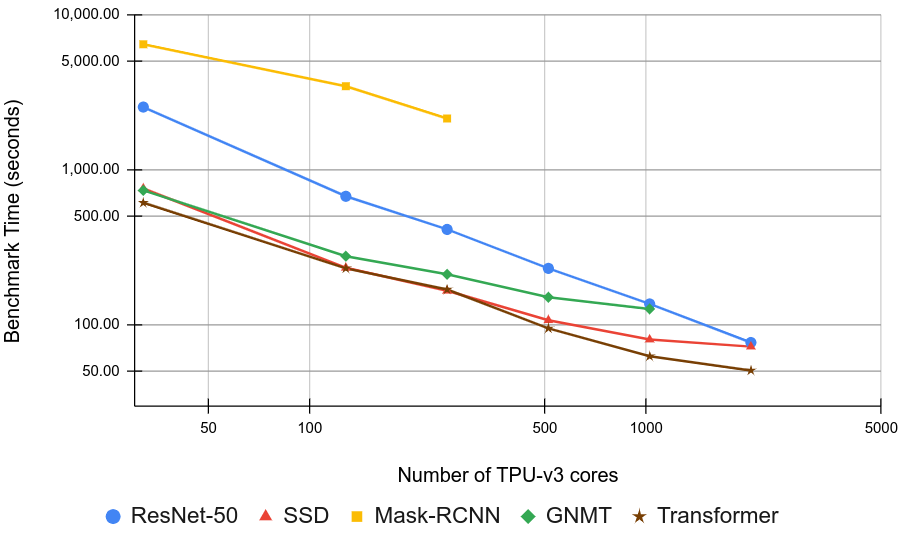}
        \caption{MLPerf-0.6 benchmark seconds.}
        \label{fig:benchmark}
    \end{minipage}\hfill
    \begin{minipage}{0.475\textwidth}
        \centering
        \includegraphics[width=1.0\textwidth]{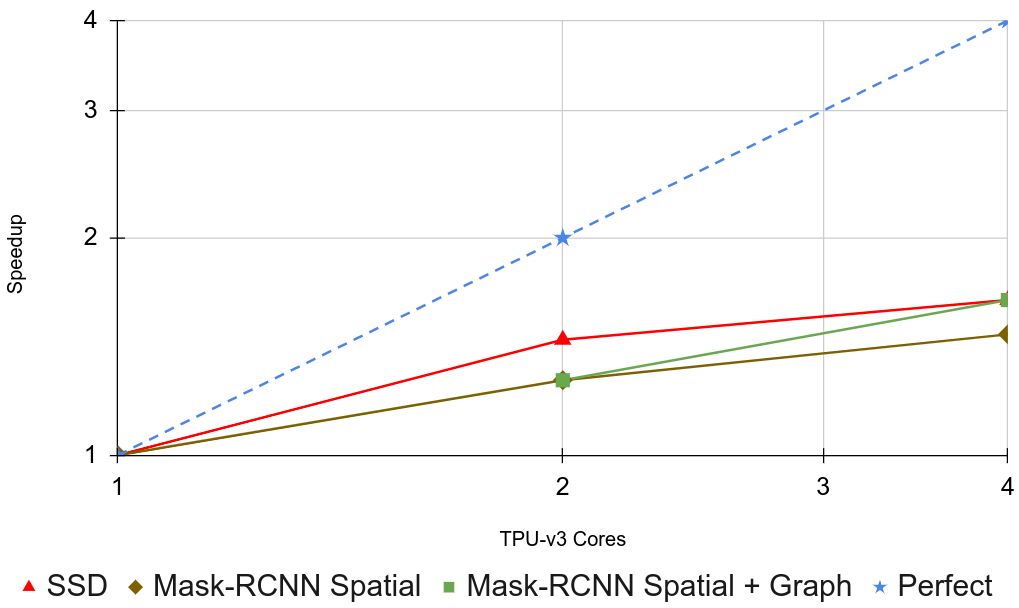}
        \caption{Speedup with model parallelism}
        \label{fig:ssd_speedup}
    \end{minipage}
\end{figure}

Figure~\ref{fig:batch_size} shows the batch sizes used in the Google
MLPerf-0.6 submissions. Note, with the exception of ResNet-50, in all
other MLPerf-0.6 models batch size only increases two times or
less. In the absence of batch parallelism, it is challenging to scale
ML workloads to a large number of accelerator cores. In addition, we
find the number of epochs to converge the model to target accuracy
increases for larger batch sizes. A comparison number of epochs to
converge vs batch size for the MLPerf modes is presented in
Figure~\ref{fig:epochs}. For example, in SSD, we need 22\% more epochs
to reach target accuracy or mAP 0.23 for SSD when increasing batch
size from 256 to 1024 and an additional 27\% more epochs at batch size
2048. Figure~\ref{fig:benchmark} presents completion times for the
five MLPerf benchmarks. In ResNet-50, GNMT and transformer we use data
parallelism, while in SSD and Mask-RCNN use both data and model
parallelism to achieve the largest scale.  With the SSD model, we
achieve a speedup of 1.6x on 4 TPU accelerator cores with
model-parallelism (Figure~\ref{fig:ssd_speedup}), enabling scaling to
2048 TPU cores. With Mask-RCNN on 128 and 256 cores, model parallelism 
is enabled across 2 and 4 cores, respectively. Speedup from model 
parallelism in Mask-RCNN is also shown in Figure~\ref{fig:ssd_speedup}.

Although the MLPerf benchmarks are batch limited, the techniques
presented in this paper enable strong scaling to 2048 TPU-v3
cores. The Google MLPerf-0.6 submissions report {\bf record
  performance} for the ResNet-50, SSD and Transformer benchmarks in
closed division category.

\section{Future Work}
Given that MLPerf is a recent benchmark suite (less than 2 years old)
and the Google TPU is still a relatively new hardware accelerator, we
believe there is significant work in this space. MLPerf will continue
to evolve and grow as a benchmark to reflect state-of-the-art in the
industry. There will still be significant work to understand large
scale models using TPU-v3 Pods by refining model parallelism
techniques and continuing to leverage compiler based optimizers such
as XLA.

MLPerf will continue to see significant evolution in models and
datasets. While a recommendation task, such as Neural Collaborative
Filtering (NCF), was absent from MLPerf-0.6, there is ongoing work to
bring a recommendation model into the MLPerf suite. Furthermore, a
speech model and dataset, such as speech-to-text, is a likely future
addition to MLPerf. We look forward to showing TPU's scalability on an
even more diverse set of models in the future.

\bibliography{bibliography}{}

\begin{thebibliography}{20}
\providecommand{\natexlab}[1]{#1}
\providecommand{\url}[1]{\texttt{#1}}
\expandafter\ifx\csname urlstyle\endcsname\relax
  \providecommand{\doi}[1]{doi: #1}\else
  \providecommand{\doi}{doi: \begingroup \urlstyle{rm}\Url}\fi

\bibitem[bf1()]{bf16}
Using bfloat16 with tensorflow models.
\newblock
  \url{https://cloud.google.com/tpu/docs/bfloat16#performance_and_memory_usage_advantages}.

\bibitem[mlp()]{mlperf}
Mlperf: Fair and useful benchmarks for measuring training and inference
  performance of ml hardware, software, and services.
\newblock \url{http://mlperf.org}.

\bibitem[tpc()]{tpch}
Tpc-h benchmark suite.
\newblock \url{http://tpc.org/tpch}.

\bibitem[wmt()]{wmt_17}
Second conference on machine translation.
\newblock \url{http://statmt.org/wmt17}.

\bibitem[xla()]{xla}
Xla: Optimizing compiler for tensorflow.
\newblock \url{https://www.tensorflow.org/xla}.

\bibitem[Abadi et~al.(2015)Abadi, Agarwal, Barham, Brevdo, Chen, Citro,
  Corrado, Davis, Dean, Devin, Ghemawat, Goodfellow, Harp, Irving, Isard, Jia,
  Jozefowicz, Kaiser, Kudlur, Levenberg, Man\'{e}, Monga, Moore, Murray, Olah,
  Schuster, Shlens, Steiner, Sutskever, Talwar, Tucker, Vanhoucke, Vasudevan,
  Vi\'{e}gas, Vinyals, Warden, Wattenberg, Wicke, Yu, and
  Zheng]{tensorflow2015-whitepaper}
Mart\'{\i}n Abadi, Ashish Agarwal, Paul Barham, Eugene Brevdo, Zhifeng Chen,
  Craig Citro, Greg~S. Corrado, Andy Davis, Jeffrey Dean, Matthieu Devin,
  Sanjay Ghemawat, Ian Goodfellow, Andrew Harp, Geoffrey Irving, Michael Isard,
  Yangqing Jia, Rafal Jozefowicz, Lukasz Kaiser, Manjunath Kudlur, Josh
  Levenberg, Dandelion Man\'{e}, Rajat Monga, Sherry Moore, Derek Murray, Chris
  Olah, Mike Schuster, Jonathon Shlens, Benoit Steiner, Ilya Sutskever, Kunal
  Talwar, Paul Tucker, Vincent Vanhoucke, Vijay Vasudevan, Fernanda Vi\'{e}gas,
  Oriol Vinyals, Pete Warden, Martin Wattenberg, Martin Wicke, Yuan Yu, and
  Xiaoqiang Zheng.
\newblock {TensorFlow}: Large-scale machine learning on heterogeneous systems,
  2015.
\newblock URL \url{https://www.tensorflow.org/}.
\newblock Software available from tensorflow.org.

\bibitem[Dean et~al.(2012)Dean, Corrado, Monga, Chen, Devin, Mao, Senior,
  Tucker, Yang, Le, et~al.]{dean2012large}
Jeffrey Dean, Greg Corrado, Rajat Monga, Kai Chen, Matthieu Devin, Mark Mao,
  Andrew Senior, Paul Tucker, Ke~Yang, Quoc~V Le, et~al.
\newblock Large scale distributed deep networks.
\newblock In \emph{Advances in neural information processing systems}, pages
  1223--1231, 2012.

\bibitem[Dixit(1991)]{DBLP:journals/pc/Dixit91}
Kaivalya~M. Dixit.
\newblock The {SPEC} benchmarks.
\newblock \emph{Parallel Computing}, 17\penalty0 (10-11):\penalty0 1195--1209,
  1991.
\newblock \doi{10.1016/S0167-8191(05)80033-X}.
\newblock URL \url{https://doi.org/10.1016/S0167-8191(05)80033-X}.

\bibitem[Goyal et~al.(2017)Goyal, Doll{\'{a}}r, Girshick, Noordhuis,
  Wesolowski, Kyrola, Tulloch, Jia, and He]{DBLP:journals/corr/GoyalDGNWKTJH17}
Priya Goyal, Piotr Doll{\'{a}}r, Ross~B. Girshick, Pieter Noordhuis, Lukasz
  Wesolowski, Aapo Kyrola, Andrew Tulloch, Yangqing Jia, and Kaiming He.
\newblock Accurate, large minibatch {SGD:} training imagenet in 1 hour.
\newblock \emph{CoRR}, abs/1706.02677, 2017.
\newblock URL \url{http://arxiv.org/abs/1706.02677}.

\bibitem[He et~al.(2015)He, Zhang, Ren, and Sun]{DBLP:journals/corr/HeZRS15}
Kaiming He, Xiangyu Zhang, Shaoqing Ren, and Jian Sun.
\newblock Deep residual learning for image recognition.
\newblock \emph{CoRR}, abs/1512.03385, 2015.
\newblock URL \url{http://arxiv.org/abs/1512.03385}.

\bibitem[He et~al.(2017)He, Gkioxari, Doll{\'{a}}r, and
  Girshick]{DBLP:journals/corr/HeGDG17}
Kaiming He, Georgia Gkioxari, Piotr Doll{\'{a}}r, and Ross~B. Girshick.
\newblock Mask {R-CNN}.
\newblock \emph{CoRR}, abs/1703.06870, 2017.
\newblock URL \url{http://arxiv.org/abs/1703.06870}.

\bibitem[Jouppi et~al.(2017)Jouppi, Young, Patil, Patterson, Agrawal, Bajwa,
  Bates, Bhatia, Boden, Borchers, Boyle, Cantin, Chao, Clark, Coriell, Daley,
  Dau, Dean, Gelb, Ghaemmaghami, Gottipati, Gulland, Hagmann, Ho, Hogberg, Hu,
  Hundt, Hurt, Ibarz, Jaffey, Jaworski, Kaplan, Khaitan, Koch, Kumar, Lacy,
  Laudon, Law, Le, Leary, Liu, Lucke, Lundin, MacKean, Maggiore, Mahony,
  Miller, Nagarajan, Narayanaswami, Ni, Nix, Norrie, Omernick, Penukonda,
  Phelps, Ross, Salek, Samadiani, Severn, Sizikov, Snelham, Souter, Steinberg,
  Swing, Tan, Thorson, Tian, Toma, Tuttle, Vasudevan, Walter, Wang, Wilcox, and
  Yoon]{Jouppi_17}
Norman~P. Jouppi, Cliff Young, Nishant Patil, David~A. Patterson, Gaurav
  Agrawal, Raminder Bajwa, Sarah Bates, Suresh Bhatia, Nan Boden, Al~Borchers,
  Rick Boyle, Pierre{-}luc Cantin, Clifford Chao, Chris Clark, Jeremy Coriell,
  Mike Daley, Matt Dau, Jeffrey Dean, Ben Gelb, Tara~Vazir Ghaemmaghami,
  Rajendra Gottipati, William Gulland, Robert Hagmann, Richard~C. Ho, Doug
  Hogberg, John Hu, Robert Hundt, Dan Hurt, Julian Ibarz, Aaron Jaffey, Alek
  Jaworski, Alexander Kaplan, Harshit Khaitan, Andy Koch, Naveen Kumar, Steve
  Lacy, James Laudon, James Law, Diemthu Le, Chris Leary, Zhuyuan Liu, Kyle
  Lucke, Alan Lundin, Gordon MacKean, Adriana Maggiore, Maire Mahony, Kieran
  Miller, Rahul Nagarajan, Ravi Narayanaswami, Ray Ni, Kathy Nix, Thomas
  Norrie, Mark Omernick, Narayana Penukonda, Andy Phelps, Jonathan Ross, Amir
  Salek, Emad Samadiani, Chris Severn, Gregory Sizikov, Matthew Snelham, Jed
  Souter, Dan Steinberg, Andy Swing, Mercedes Tan, Gregory Thorson, Bo~Tian,
  Horia Toma, Erick Tuttle, Vijay Vasudevan, Richard Walter, Walter Wang, Eric
  Wilcox, and Doe~Hyun Yoon.
\newblock In-datacenter performance analysis of a tensor processing unit.
\newblock In \emph{Proceedings of ISCA'17}, 2017.
\newblock URL \url{http://arxiv.org/abs/1704.04760}.

\bibitem[Keskar et~al.(2016)Keskar, Mudigere, Nocedal, Smelyanskiy, and
  Tang]{DBLP:journals/corr/KeskarMNST16}
Nitish~Shirish Keskar, Dheevatsa Mudigere, Jorge Nocedal, Mikhail Smelyanskiy,
  and Ping Tak~Peter Tang.
\newblock On large-batch training for deep learning: Generalization gap and
  sharp minima.
\newblock \emph{CoRR}, abs/1609.04836, 2016.
\newblock URL \url{http://arxiv.org/abs/1609.04836}.

\bibitem[Lin et~al.(2014)Lin, Maire, Belongie, Bourdev, Girshick, Hays, Perona,
  Ramanan, Doll{\'{a}}r, and Zitnick]{microsoft_coco}
Tsung{-}Yi Lin, Michael Maire, Serge~J. Belongie, Lubomir~D. Bourdev, Ross~B.
  Girshick, James Hays, Pietro Perona, Deva Ramanan, Piotr Doll{\'{a}}r, and
  C.~Lawrence Zitnick.
\newblock Microsoft {COCO:} common objects in context.
\newblock \emph{CoRR}, abs/1405.0312, 2014.
\newblock URL \url{http://arxiv.org/abs/1405.0312}.

\bibitem[Liu et~al.(2015)Liu, Anguelov, Erhan, Szegedy, Reed, Fu, and
  Berg]{ssd_15}
Wei Liu, Dragomir Anguelov, Dumitru Erhan, Christian Szegedy, Scott~E. Reed,
  Cheng{-}Yang Fu, and Alexander~C. Berg.
\newblock {SSD:} single shot multibox detector.
\newblock \emph{CoRR}, abs/1512.02325, 2015.
\newblock URL \url{http://arxiv.org/abs/1512.02325}.

\bibitem[Russakovsky et~al.(2015)Russakovsky, Deng, Su, Krause, Satheesh, Ma,
  Huang, Karpathy, Khosla, Bernstein, Berg, and Fei-Fei]{ILSVRC15}
Olga Russakovsky, Jia Deng, Hao Su, Jonathan Krause, Sanjeev Satheesh, Sean Ma,
  Zhiheng Huang, Andrej Karpathy, Aditya Khosla, Michael Bernstein,
  Alexander~C. Berg, and Li~Fei-Fei.
\newblock {ImageNet Large Scale Visual Recognition Challenge}.
\newblock \emph{International Journal of Computer Vision (IJCV)}, 115\penalty0
  (3):\penalty0 211--252, 2015.
\newblock \doi{10.1007/s11263-015-0816-y}.

\bibitem[Vaswani et~al.(2017)Vaswani, Shazeer, Parmar, Uszkoreit, Jones, Gomez,
  Kaiser, and Polosukhin]{DBLP:journals/corr/VaswaniSPUJGKP17}
Ashish Vaswani, Noam Shazeer, Niki Parmar, Jakob Uszkoreit, Llion Jones,
  Aidan~N. Gomez, Lukasz Kaiser, and Illia Polosukhin.
\newblock Attention is all you need.
\newblock \emph{CoRR}, abs/1706.03762, 2017.
\newblock URL \url{http://arxiv.org/abs/1706.03762}.

\bibitem[Wu et~al.(2016)Wu, Schuster, Chen, Le, Norouzi, Macherey, Krikun, Cao,
  Gao, Macherey, Klingner, Shah, Johnson, Liu, Kaiser, Gouws, Kato, Kudo,
  Kazawa, Stevens, Kurian, Patil, Wang, Young, Smith, Riesa, Rudnick, Vinyals,
  Corrado, Hughes, and Dean]{GNMT_16}
Yonghui Wu, Mike Schuster, Zhifeng Chen, Quoc~V. Le, Mohammad Norouzi, Wolfgang
  Macherey, Maxim Krikun, Yuan Cao, Qin Gao, Klaus Macherey, Jeff Klingner,
  Apurva Shah, Melvin Johnson, Xiaobing Liu, Lukasz Kaiser, Stephan Gouws,
  Yoshikiyo Kato, Taku Kudo, Hideto Kazawa, Keith Stevens, George Kurian,
  Nishant Patil, Wei Wang, Cliff Young, Jason Smith, Jason Riesa, Alex Rudnick,
  Oriol Vinyals, Greg Corrado, Macduff Hughes, and Jeffrey Dean.
\newblock Google's neural machine translation system: Bridging the gap between
  human and machine translation.
\newblock \emph{CoRR}, abs/1609.08144, 2016.
\newblock URL \url{http://arxiv.org/abs/1609.08144}.

\bibitem[Ying et~al.(2018)Ying, Kumar, Chen, Wang, and Cheng]{resnet_18}
Chris Ying, Sameer Kumar, Dehao Chen, Tao Wang, and Youlong Cheng.
\newblock Image classification at supercomputer scale.
\newblock \emph{CoRR}, abs/1811.06992, 2018.
\newblock URL \url{http://arxiv.org/abs/1811.06992}.

\bibitem[You et~al.(2017)You, Gitman, and Ginsburg]{lars_32k}
Yang You, Igor Gitman, and Boris Ginsburg.
\newblock Scaling {SGD} batch size to 32k for imagenet training.
\newblock \emph{CoRR}, abs/1708.03888, 2017.
\newblock URL \url{http://arxiv.org/abs/1708.03888}.

\end{thebibliography}
\bibliographystyle{plainnat}

\end{document}